\pdfoutput=1
\documentclass[11pt]{article}

\usepackage[final]{acl}

\usepackage{times}
\usepackage{latexsym}

\usepackage[T1]{fontenc}
\usepackage[utf8]{inputenc}

\usepackage{microtype}

\usepackage{inconsolata}

\usepackage{tikz}
\usepackage{xcolor}
\usepackage{amsmath}
\usetikzlibrary{fit}
\usetikzlibrary{backgrounds}
\usetikzlibrary{shapes.geometric, arrows, positioning}
\usetikzlibrary{arrows.meta}
\usepackage{graphicx}
\usepackage{bbm}
\usepackage{bm}

\title{Hypernetworks for Perspectivist Adaptation}

\author{
 \textbf{Daniil Ignatev\textsuperscript{1}},
 \textbf{Denis Paperno\textsuperscript{1}},
 \textbf{Massimo Poesio\textsuperscript{1,2}},
\\
\\
 \textsuperscript{1}Utrecht University,
 \textsuperscript{2}Queen Mary University of London,
\\
 \small{
   \textbf{Correspondence:} \href{mailto:d.ignatev@uu.nl}{d.ignatev@uu.nl}
 }
}

\begin{document}
\maketitle
\begin{abstract}
The task of perspective-aware classification introduces a bottleneck in terms of parametric efficiency that did not get enough recognition in existing studies. In this article, we aim to address this issue by applying an existing architecture, the \texttt{hypernetwork+adapters} combination, to perspectivist classification. Ultimately, we arrive at a solution that can compete with specialized models in adopting user perspectives on hate speech and toxicity detection, while also making use of considerably fewer parameters. Our solution is architecture-agnostic and can be applied to a wide range of base models out of the box.
\end{abstract}

\section{Introduction}

In the recent years, perspective-aware approach to subjective linguistic tasks has been gaining prominence in NLP. This approach suggests that for tasks that involve subjectivity, dataset designers should collect multiple labels from different annotators for each data instance; these labels need to be retained and used in model training \cite{plank-2022-problem,Cabitza_Campagner_Basile_2023,fleisig-etal-2024-perspectivist}. This policy is warranted in tasks like hate speech detection, where multiple labels assigned by annotators with diverse backgrounds can be equally applicable. This notion contrasts the popular practice of designating a single supposedly correct label for each bit of data while discarding conflicting annotator judgments.

Likewise, NLP researchers have argued that subjective tasks require perspective-aware machine learning methods, i.e., methods that can capture diverse opinions based on unaggregated labels \cite{akhtar2021opinionsmatterperspectiveawaremodels}. We are particularly interested in the paradigm of \textit{strong perspectivism} \cite{Cabitza_Campagner_Basile_2023}. The latter suggests that much like model personalization, separate models or representations should be trained on labels produced by individual annotators. In practice, this entails one of the following: (1) training a full language model checkpoint on each annotator's labels; (2) when using PEFT methods \cite{pmlr-v97-houlsby19a}, training a separate adaptation component for each perspective; (3) using a specialized language model architecture.\footnote{Here, we do not consider few-shot and zero-shot learning, as they do not fall under the training category.}
We argue that both the first and the second strategy require a prohibitive amount of trainable parameters that increases with model size and number of modeling targets (see discussion in Section \ref{sec:discussion}). To tackle this problem, we make use of an architecture that employs a small trainable module and adapts a model to diverse perspectives while keeping most parameters frozen. 

The core of our solution is the \texttt{hypernetwork+adapters} combination. A hypernetwork is a neural architecture in which a source neural network is trained to predict the weights of a target network; these weights are subsequently used in inference \cite{ha-2017-hypernetworks}. Importantly, hypernetworks can also be used to predict weights of adapters, including LoRA adapters \cite{pmlr-v97-houlsby19a,hu2022lora}, rather than weights of entire models. Adapters can be defined as small trainable modules designed for parameter-efficient tuning of NLP models including large language models; compatibility with adapters makes hypernetworks suitable for the same task \cite{karimi-mahabadi-etal-2021-parameter,he2022hyperprompt,phang2023hypertuning}. We provide more details on this architecture in Section \ref{sec:method}. The novel aspect of our work is that we attempt to repurpose the \texttt{hypernetwork+adapters} combination for perspectivist modeling and, by extension, model personalization. We posit its strengths, i.e., the reduction of trainable parameters, as a way to address the mentioned bottleneck issue. To our knowledge, this direction has not been thoroughly explored in previous studies.

% The novel aspect our approach is that hypernetworks have not been applied to perspectivist modeling — a gap this paper attempts to bridge. We see a lot of promise in this approach for the following reasons. Adapter-based architecture naturally fits into the perspectivist paradigm, since fine-tuning an entire language model for each annotator perspective is computationally prohibitive; meanwhile, adapters leverage a mere fraction of the original parameters. However, training an adapter for every perspective may also prove unfeasible, especially in the case of large language models (LLMs) that require larger adapter layers. In contrast, adapter-modeling hypernetworks have a constant size and, as a result, can adapt arbitrarily large models to a multitude of perspectives. 
% % However, the applicability of this architecture calls for thorough testing; we address this question in our experiments (Section TODO ...).

Our article makes the following contributions.
First, we consider whether \texttt{hypernetwork+adapters} setting is generally applicable to annotator-aware text classification. Our results suggest that hypernetworks are well fit for this task, albeit not unconditionally.

Second, in our experiments,\footnote{Our codebase has been made publicly accessible at \href{https://github.com/ruthenian8/Hypernets}{https://github.com/ruthenian8/Hypernets}} we show that our hypernetwork-based architecture performs on par with recent perspectivist model architectures — particularly, Annotator-Aware Representations for Texts \cite{mokhberian-etal-2024-capturing}, and Annotator Embeddings \cite{deng2023annotate}. At the same time, we note the limitations of the proposed architecture and attempt to address them.

Third, we demonstrate that the proposed architecture offers a better trade-off in terms of parameter efficiency than the baseline models. At the same time, it also demonstrates a greater degree of versatility: since the weights of the base model are not being affected, there is no risk of forgetting \cite{kirkpatrick2017overcoming} or degradation of the base model's fundamental capabilities. These properties of our method make us believe that it has promise in modeling subjective tasks.
% hypernetworks do not affect the base model's internals much like LoRA adapters, they can be trivially plugged into a broad range of models, including LLMs. We also release our implementation of an adapter-modeling hypernetwork based on TODO ... 's description, but programmed independently.

\section{Related work}

\noindent \paragraph{Data perspectivism:} the idea of preserving pluralistic annotations in datasets has been discussed for a considerable time \cite{artstein-poesio-2008-survey} and has been gaining increasing prominence across various AI domains \cite{10.5555/3563572.3563588,10.1145/3544548.3580645,huang-etal-2023-incorporating}. \citet{frenda2024perspectivist} provides a detailed survey of such perspectivist datasets and methods, reflecting a paradigm shift towards treating annotator disagreement not as noise but as a source of valuable signal about human variation.

\noindent \paragraph{Perspectivist learning:} modeling human variation in labeling based on annotator perspectives has been advocated in several recent studies \cite{plank-2022-problem,Cabitza_Campagner_Basile_2023}. In terms of modeling techniques, various recipes have been explored, including trainable crowd layers \cite{10.5555/3504035.3504232}; trainable tokens \cite{efb1f5ad5715cdff23dc5841ae9db853d901dd4a}; multi-task classification with annotator-specific heads \cite{davani-etal-2022-dealing}, and others. Recently, studies have focused on active learning \cite{3ecfd550ac7eb0d375c8432dc1e12c453110342e,wang-plank-2023-actor} and few-shot perspective modeling \cite{8d687f8edca283308d024fa3016cc18ec4a21c1e,sorensen2025valueprofilesencodinghuman} as two ways to deal with data sparsity. 

Strong perspectivism brings together perspectivist learning and model personalization, which is why recent methods from the latter domain are also relevant to our research. In particular, studies by \citet{tan2024democratizing} and \citet{clarke2024peft} show that adapters and LoRA adapters can rival full model finetuning on the LLM personalization task, even when the number of trainable parameters is reduced to a fraction of the original model size.

Finally, in our study, we are particularly interested in recent works that optimize encoder-based classifiers against pluralistic labels and make use of annotator embeddings \cite{deng2023annotate,mokhberian-etal-2024-capturing}. In this work, we consider both of these as our baselines.

\noindent \paragraph{Hypernetworks:} 
The idea of hypernetworks has its roots in an earlier concept of weight generators \cite{10.1007/11550907_61} and aims to constrain the search space when modeling complex objectives through searching in the limited weight space. Practically, this means generating the weights of a target model by means of a separate generator model (hypernetwork). 

The title paper, published in 2017 \cite{ha-2017-hypernetworks}, had two additional objectives: reduction of trainable parameters (1) and regularized training (2). The study proposed an implementation of two network types: a CNN network generating weights for another CNN network and an RNN network producing weights for a target RNN. Their results showed that the system achieves competitive performance and reduces the trainable parameter count, effectively fulfilling its purpose. The authors' assumptions were further scrutinized in follow-up works. As an example, a study by \citet{soydaner2020hyper} applied hypernetworks to convolutional autoencoder models and showed that the weights of an autoencoder can be closely approximated by a much smaller hypernetwork, resulting in significant reduction of trainable parameters. 

Moreover, the ML community recognized early the potential of hypernetworks in multitask learning. For instance, a study by \citet{tay2021hypergrid} adapts a transformer model to a multitude of tasks by predicting task-specific weights for the model's feed-forward layer using a hypernetwork. This adaptation strategy is of particular interest to us due to \citet{davani-etal-2022-dealing}'s success in learning perspectives through multi-task classification. 

The above architectures leveraged hypernetworks to predict the model's own weights. However, the parameter-efficient finetuning paradigm brought about an alternative approach, which is to infer the weights of small trainable submodules while leaving the weights of the main model intact. Adapters are a specific instance of these submodules \cite{pmlr-v97-houlsby19a}; modeling them with hypernetworks has led to impactful results in several machine learning applications. Specifically, in image generation, hypernetworks have been used to personalize diffusion models achieving a considerable speed-up compared to other methods \cite{Ruiz2023HyperDreamBoothHF}. In a different vein, in speech recognition, \citet{muller-eberstein-etal-2024-hypernetworks} employed hypernetworks to adapt ASR models to individual speakers with atypical speech patterns. Their hypernetwork obtains a 75\% relative reduction in word error rate using only 0.1\% of the model parameters. 

In NLP, hypernetworks have been utilized for multi-task and multilingual adaptation of larger transformer models, such as T5 \cite{raffel2020exploring}. \citet{ustun-etal-2022-hyper} propose a single hypernetwork that produces adapter weights for multiple languages and tasks simultaneously, eliminating the need for separate language-specific task adapters. \citet{karimi-mahabadi-etal-2021-parameter} and \citet{phang2023hypertuning} leverage a shared hypernetwork to effectively train adapter modules for a range of NLP tasks. Finally, in 2025, \citet{charakorn2025text} used hypernetworks to generate task-sepecific LoRA parameters for LLMs on the fly given a textual task description, thus enabling zero-shot adaptation; their results show that hypernetworks are combinable with LLMs, while their approach may also be repurposed for user personalization in the future.

% \cite{rizzi2025many}: thesis, perspectivism, hate speech

% The solutions that have been proposed can be divided into several categories: learning from per-item label distributions, aka soft labels \cite{peterson2019human,Uma_Fornaciari_Hovy_Paun_Plank_Poesio_2020}, and learning annotator-specific labels, whether through multi-task learning \cite{davani-etal-2022-dealing} or through training separate models or representations for each annotator or annotator group \cite{mokhberian-etal-2024-capturing}. We also refer to the latter paradigm as perspectivist modeling.

\section{Method}
\label{sec:method}

\subsection{Architecture}
\label{sec:architecture}

\begin{figure}
    \centering
    \resizebox{0.5\textwidth}{!}{
    \begin{tikzpicture}[
      node distance=1mm and 2mm,
      every node/.style={font=\small},
      inviz/.style={rectangle, rounded corners, minimum width=1cm, minimum height=6mm},
      embed/.style={rectangle, draw, rounded corners, fill=blue!10, minimum width=1cm, minimum height=6mm},
      proc/.style={rectangle, draw, rounded corners, fill=green!10, minimum width=1cm, minimum height=6mm},
      head/.style={rectangle, draw, rounded corners, fill=orange!10, minimum width=1cm, minimum height=6mm},
      mat/.style={draw, fill=purple!10, minimum width=1cm, minimum height=6mm},
      arrow/.style={-{Stealth[]}, thick},
    ]
    
    % Inputs
    \node[minimum width=1cm] (annotator) {ID$_{\text{ann}}$};
    \node[inviz, draw=none, below=of annotator] (inviz1) {};
    \node[below=of inviz1, minimum width=1cm] (layerid) {ID$_{\text{layer}}$};
    
    % Embeddings
    \node[embed, right=of annotator, xshift=3mm] (e1) {E$_{\text{ann}}$};
    \node[inviz, draw=none, right=of inviz1, xshift=6mm] (inviz2) {};
    \node[embed, right=of layerid, xshift=6mm] (e2) {E$_{\text{layer}}$};
    
    % Concat
    \node[proc, right=of inviz2, xshift=1mm] (concat) {$Ea|| E_l$};
    
    \node[inviz, draw=none, right=of concat] (lin2) {};
    
    % Heads for A & B
    \node[head, above=of lin2] (headA) {Lin$_{A}$};
    \node[head, below=of lin2] (headB) {Lin$_{B}$};
    
    % Matrices
    \node[mat, trapezium, trapezium angle=-70, right=of lin2, xshift=3mm, yshift=10] (A) {LoRA$A_{ij}$};
    \node[mat, trapezium, trapezium angle=70, right=of lin2, xshift=3mm, yshift=-10] (B) {LoRA$B_{ij}$};
    \node[above=of A, yshift=7] (input) {Text input};
    \node[below=of B, yshift=-7] (output) {Label};
    
    % Arrows
    \draw[arrow] (annotator) -- (e1);
    \draw[arrow] (layerid) -- (e2);
    
    \draw[arrow] (e1) -- (concat);
    \draw[arrow] (e2) -- (concat);
    
    \draw[arrow] (concat) -- (headA);
    \draw[arrow] (concat) -- (headB);
    
    \draw[arrow] (headA) -- (A);
    \draw[arrow] (headB) -- (B);
    \draw[arrow] (input) -- (A);
    \draw[arrow] (B) -- (output);
    
    % Optional legend / grouping
    \begin{scope}[on background layer]
       \node[draw, dashed, inner sep=6pt, fit={(annotator) (inviz1) (layerid)}, label=above:HN inputs] {};
      \node[draw, dashed, inner sep=6pt, fit={(e1) (e2) (concat) (lin2) (headA) (headB)}, label=above:Hypernetwork] {};
      \node[draw, rounded corners, fill=gray!20, inner sep=5pt, fit={(A) (B)}] (BM) {};
      \node[rotate=-90, right=of BM, xshift=-15] {Layer $j$};
    \end{scope}
    
    \end{tikzpicture}
}
    \caption{Data flow within our hypernetwork implementation: the annotator id and the layer id are embedded to adapt the target layer $l_j$ in the base model.}
    \label{fig:schema}
\end{figure}
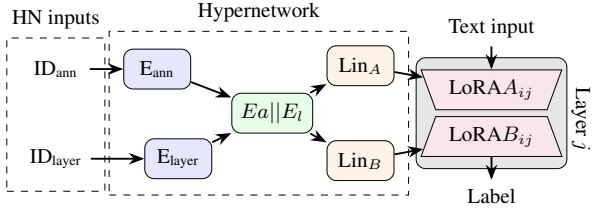

The proposed architecture aims to tune a base model to each annotator's perspective; it implements an adapter-modelling hypernetwork in a fashion similar to \citealt{phang2023hypertuning} and \citealt{muller-eberstein-etal-2024-hypernetworks}, as it specifically makes use of low-rank adapters \cite{hu2022lora}. Generally, all adapter variations enable parameter-efficient fine-tuning of large models by freezing and patching the base model's pre-trained weights $W_j$ within a layer $l_j$ with a smaller trainable layer $Ada_j$. Low-rank adapters, in particular, take this reduction of updatable parameters one step further by decomposing $Ada_j$ into two low-rank matrices, $A_j$ and $B_j$; ultimately, they decrease the parameter count even more with little performance impact. This principle can thus be illustrated with the following formula:

\begin{flalign}
    W_j' = W_j + B_jA_j
\end{flalign}

\noindent where $A_j$ and $B_j$ are low-rank trainable matrices. We find this type of adapters preferable for use in combination with hypernetworks, as it narrows down the solution space for the hypernetwork component.

In our implementation, when modeling the perspective $Ann_i$, we use the hypernetwork $H$ to predict $A_{ij}$ and $B_{ij}$ for every layer $l_j$. Hence, we need to condition our hypernetwork on two relevant variables: information on the target annotator $Ann_i$ and on the target layer $l_j$.

\begin{flalign}
    A_{ij}, B_{ij} = H(Ann_i, l_j)
\end{flalign}

For demonstration purposes, we only consider unique annotator IDs as annotator information. However, other relevant variables can also be straightforwardly integrated. We leave it to further research to determine whether using more complex representations based on sociodemographic variables or prior annotations shows better efficiency. 

Likewise, target layer information is supplied through numeric layer identifiers. In the hypernetwork module, both types of identifiers are embedded, concatenated, and jointly passed to two prediction heads ($Lin_A, Lin_B$), which then infer the matrices A and B; this flow is illustrated in Figure \ref{fig:schema}.

We add the hypernetwork component on top of PEFT's architecture-agnostic LoRA implementation \cite{peft} with the aim of making our method compatible with a wide range of base models. To mitigate possible generalization issues, we use GeLU activation \cite{hendrycks2016gaussian} and dropout probability of 0.25 in the hypernetwork module. We also follow the suggestion of \citealt{muller-eberstein-etal-2024-hypernetworks} by initializing $Lin_B$ with zeros and $Lin_A$ with values close to zero, ensuring smooth updates at the initial stages of training.

\subsection{Data}

In this study, we purposefully pick 4 datasets that permit us to compare our architecture against existing algorithms for perspectivist learning. Two of these datasets, HS-Brexit and MD-Agreement, were included in the shared task on Learning With Disagreements (LeWiDi, \citealt{leonardelli-etal-2023-semeval}). This section gives an overview of all data we used.

\paragraph{The Multi-Domain Agreement dataset ($\mathcal{D}_{\textsl{MDA}}$):} This dataset by \citet{leonardelli-etal-2021-agreeing} addresses the task of offensive language detection. MD-Agreement comprises 9,814 English tweets from three distinct domains: the Black Lives Matter movement, the 2020 Election, and the COVID-19 pandemic. Each tweet was annotated by at least 5 Mechanical Turk workers out of a total pool of 334.

\paragraph{English Perspectivist Irony Corpus ($\mathcal{D}_{\textsl{EPIC}}$):}
Introduced by \citet{frenda-etal-2023-epic}, the corpus consists of 3,000 Post-Reply pairs collected from Twitter and Reddit. The data was sourced from five English-speaking countries: Australia, India, Ireland, the United Kingdom, and the United States. A total of 74 annotators, balanced by gender and nationality, participated in the annotation task, with around 15 raters per nationality. Each annotator labeled approximately 200 instances, resulting in a corpus with 14,172 annotations and a median of 5 annotations per instance.

\paragraph{The Racial Bias Toxicity Detection Corpus ($\mathcal{D}_{\textsl{RB}}$):}
\citet{sap-etal-2019-risk} investigated the interaction between annotators' biases and their perceptions of toxicity reaching positive conclusions. The authors recruited 819 Amazon Mechanical Turk workers to annotate tweets for two variables: whether a tweet was (a) personally offensive to them and (b) potentially offensive to others. As in \citealt{mokhberian-etal-2024-capturing}, our experiments focus on (a).

\paragraph{Hate Speech Brexit ($\mathcal{D}_{\textsl{HS-Brexit}}$):} \citet{akhtar2021opinionsmatterperspectiveawaremodels} had 6 experiment participants label the same set of 1120 tweets related to Brexit. Each tweet was annotated for several categories including Hate Speech,
Aggressiveness, Offensiveness, and Stereotype. In our study, we focus on Hate Speech annotations. 

\begin{table}[ht!]
\small
\centering
\begin{tabular}{lccccc}
\multicolumn{1}{l|}{\textit{Dataset}} & \textit{Train} & \textit{Dev} & \textit{Test} & \textit{\#A} & \textit{\#E/\#A}
\\ \hline
\multicolumn{1}{l|}{$\mathcal{D}_{\textsl{MDA}}$} & $27k$ & $13k$ & $13k$ & 334 & 160 \\
\multicolumn{1}{l|}{$\mathcal{D}_{\textsl{EPIC}}$} & $7.1 k$ & $3.5k$ & $3.5k$ & 74 & 191 \\
\multicolumn{1}{l|}{$\mathcal{D}_{\textsl{RB}}$} & $6.1k$ & $2.7k$ & $2.8k$ & 819 & 14 \\
\multicolumn{1}{l|}{$\mathcal{D}_{\textsl{HS-Brexit}}$} & $3.3k$ & $1.6k$ & $1.6k$ & 6 & 1120
\\ \hline
\end{tabular}
\caption{Post-split statistics of datasets used in the experiments. \textit{\#A} stands for the number of workers; \textit{\#E/\#A} denotes the mean number of annotated items per worker.}
\label{tab:datasets}
\end{table}

For our experiments, we reproduce the dataset splitting procedure from \citealt{mokhberian-etal-2024-capturing}. Specifically, we split the data into partitions of 50, 25, and 25\% (train, dev, and test sets respectively) stratified with respect to item-level disagreements. Items from the dev and test sets annotated by an annotator not present in the training set are merged into the training data. This splitting procedure is repeated using 10 different random seeds. We report the post-preprocessing statistics for the four datasets in Table \ref{tab:datasets}.

\section{Experiments}

\subsection{Baselines}

We test our architecture against approaches introduced in \citealt{deng2023annotate} (AE) and \citealt{mokhberian-etal-2024-capturing} (AART). Both of these build on generic transformer models and handle diverse perspectives by integrating them directly into the architecture. To that end, they make use of specialized annotator representations.

\noindent \paragraph{AART:} The AART architecture combines text embeddings from pretrained transformer models with learned annotator embeddings. Formally, given a text item $x_i$ and annotator $a_j$, the combined embedding is computed as:

$$
g(x_i, a_j) = e(x_i) + f(a_j)
$$

where $e(x_i)$ is the text embedding, and $f(a_j)$ is a learned annotator-specific embedding. This combined embedding is then fed into a common classification head. The model employs a multipart loss function consisting of cross-entropy loss, L2 regularization on annotator embeddings, and a contrastive loss designed to cluster similar annotator perspectives. We compare our results against AART on $\mathcal{D}_{\textsl{MDA}}$, $\mathcal{D}_{\textsl{EPIC}}$, and $\mathcal{D}_{\textsl{RB}}$.

\noindent \paragraph{Annotator Embeddings (AE):} This approach explicitly captures annotator-specific biases and annotation tendencies using two types of embeddings: annotator embeddings ($E_a$) and annotation embeddings ($E_n$). These embeddings are combined with the original text embeddings via weighted summation:

$$
E_{\text{combined}} = E_{\text{[CLS]}} + \alpha_n E_n + \alpha_a E_a
$$

where $\alpha_n$ and $\alpha_a$ are learnable weights computed based on the interaction between the sentence embedding and annotator-specific embeddings. The resulting embedding is fed into transformer-based classification models, enhancing their ability to predict annotator-specific labels by explicitly modeling annotator idiosyncrasies. We compare the performance of our model against AE on $\mathcal{D}_{\textsl{MDA}}$ and $\mathcal{D}_{\textsl{HS-Brexit}}$.

\noindent \paragraph{Single-task baseline:} Both AART and AE compare their systems against a single-task baseline model. This baseline is a transformer classifier trained on majority-aggregated labels that predicts one label per text and effectively ignores annotator-specific nuances. This baseline mostly serves as a sanity check to evaluate the effectiveness of perspectivist modeling under the assumption that specialized architectures should handle label diversity at least somewhat better than a non-specialized model. However, on data items where most annotators agree with the majority label (e.g., 7 raters out of 10), it can show better results than a perspectivist model due to a much narrower problem space. Such data items constitute a large part of the existing perspectivist datasets, and, as a result of that, specialized models do not always surpass this baseline.

\subsection{Setup}
\label{sec:setup}

In all experiments, we use RoBERTa-base \cite{liu2019roberta} as our base model to maintain consistency with the baseline methods, as both studies include reported results for RoBERTa-base among other models. As a part of our solution, all layers except the hypernetwork get explicitly frozen. We model LoRA adapters  with a rank of 2 and $\alpha=32$, keeping adapter dimensionality to a minimum. Although increasing the number of trainable parameters is likely to lead to better results, we explore the lower performance boundary of our method that also offers the best efficiency tradeoff.

We train the hypernetwork for 5 epochs with a batch size of 100 at a learning rate of $1\mathrm{e}{-5}$ making use of Adam optimizer \cite{Kingma2014AdamAM}. We set max. sequence length to 100. For each dataset, we repeat the experiment 10 times with varying random seeds. To implement our method, we use abstractions from PEFT \cite{peft} and Transformers \cite{wolf-etal-2020-transformers}.

We obtained the above hyperparameter values by running a grid search with varying dropout probabilities (0.0, 0.1, 0.25) and learning rates ($5\mathrm{e}{-5}$, $1\mathrm{e}{-5}$, $5\mathrm{e}{-6}$) using $\mathcal{D}_{\textsl{EPIC}}$'s development set as a reference. Our conclusion is that unlike the baseline solutions without PEFT methods, our architecture shows greater sensitivity to learning rate and dropout probability. In particular, when using low dropout ($0$ or $0.1$) or a higher learning rate ($5\mathrm{e}{-5}$), the model often converges prematurely at a suboptimal point. This yields micro-f1 values of $\approx50.0$ suggesting that the model only learns the majority label due to label imbalance; in contrast, when using our ultimate parameter values (dropout$=0.25$, lr$=1\mathrm{e}{-5}$), training proceeds with gradual parameter updates and leads to better generalization ($\approx68.5$). This result demonstrates that careful hyperparameter tuning is necessary for our approach.

\subsection{Evaluation}
\label{sec:evaluation}

Since AART and AE were assessed with two different metric suites, we preserve the original evaluation protocols to allow direct comparison with the published figures.  In addition, we track the total number of trainable parameters for each model as described below.

\paragraph{Annotator‐level F1}
This metric measures the ability of a model to treat every annotator fairly, regardless of how many labels they contributed.  For each annotator $a_j$, we compute the macro‐F1 score over all items $x_i$ in the test split by comparing the gold label $y_{ij}$ to the model’s prediction $\hat y_{ij}$.  The \emph{Annotator‐level F1} is then the simple mean of these per‐annotator F1 scores. \citealt{mokhberian-etal-2024-capturing} argue that it prevents evaluation biases towards prolific annotators.

\paragraph{Global‐level F1}
To measure overall predictive quality, we pool all $(x_i,a_j)$ pairs in the test set and compute macro‐F1 on this combined set. Unlike the annotator‐level metric, this score weighs each prediction equally and inadvertently prioritizes annotators who contributed more labels.

\paragraph{Item‐level Disagreement Correlation}
We quantify how well a model reproduces the true pattern of annotator disagreement on each item.  For an item $x_i$ with annotator votes $\{y_{i1},\dots,y_{iK}\}$, the gold disagreement is

$$
    d_i = 1 - \frac{\max_{c} \bigl|\{j : y_{ij}=c\}\bigr|}{K},
$$

and the model‐predicted disagreement $\hat d_i$ is computed analogously from $\{\hat y_{ij}\}$.  We then report the Pearson correlation $\mathrm{Corr}(\{d_i\},\{\hat d_i\})$, as in AART.

\paragraph{Baseline‐specific Metrics}
AART uses exactly the three metrics above: annotator‐level F1, global‐level F1, and item‐level disagreement correlation.  We thus report all three for direct apples‐to‐apples comparison.  AE instead evaluates each annotator’s labels by computing (i) global accuracy over all $(x_i,a_j)$ pairs and (ii) global‐level F1 as defined above.  Because AE does not release per‐annotator predictions, we we are unable to use the annotator-level metrics for that baseline.

\paragraph{Trainable Parameters}
Lastly, we report the approximate number of trainable parameters for each model (our hypernetwork+adapters, AART, and AE).  For AART and AE, we calculate these numbers based on their public source code, while for our method we report the number directly. Analyzing parameter counts helps us inquire into the parameter efficiency of each solution and assess how well each architecture can scale to the growing number of annotators to model and growing embedding space.

\section{Results}

\begin{table}[ht!]
\centering
\small
\begin{tabular}{lccc}
\multicolumn{1}{l|}{\textit{Dataset}}       & \textit{Single-task } & \textit{AART} & \textit{Ours}  \\[0.7mm] \hline
\multicolumn{4}{c}{\textit{Annotator-level F1}}         \\ \hline
\multicolumn{1}{l|}{$\mathcal{D}_{\textsl{MDA}}$} & $66.80 \pm 0.7$ & $69.72 \pm 1.1$ & \bm{$70.24 \pm 0.9$} \\[0.7mm]
\multicolumn{1}{l|}{$\mathcal{D}_{\textsl{EPIC}}$} & $58.59 \pm 1.9$ & \bm{$59.67 \pm 0.9$} & $53.16 \pm 1.6$ \\[0.7mm]
\multicolumn{1}{l|}{$\mathcal{D}_{\textsl{RB}}$} & $68.61 \pm 1.5$ & $71.1 \pm 3.2$  & \bm{$73.81 \pm 2.0$} \\[0.7mm]
\hline
\multicolumn{4}{c}{\textit{Global-level F1}}       
\\ \hline
\multicolumn{1}{l|}{$\mathcal{D}_{\textsl{MDA}}$} & $71.99 \pm 0.6$ & $77.38 \pm 0.4$ & \bm{$78.11 \pm 0.2$} \\[0.7mm]
\multicolumn{1}{l|}{$\mathcal{D}_{\textsl{EPIC}}$} & $60.23 \pm 1.7$ & \bm{$66.16 \pm 1.4$} & $65.11 \pm 1.2$ \\[0.7mm]
\multicolumn{1}{l|}{$\mathcal{D}_{\textsl{RB}}$} & $71.97 \pm 1.7$ & \bm{$79.96 \pm 1.9$} & $76.17 \pm 1.8$ \\[0.7mm]
\hline
\multicolumn{4}{c}{\textit{Item-level Disagreement Correlations}}        
\\ \hline
\multicolumn{1}{l|}{$\mathcal{D}_{\textsl{MDA}}$} & \small{NA} & $0.37 \pm 0.04$ & \bm{$0.42 \pm 0.02$} \\[0.7mm]
\multicolumn{1}{l|}{$\mathcal{D}_{\textsl{EPIC}}$} & \small{NA} & $0.20 \pm 0.06$ & \bm{$0.28 \pm 0.03$} \\[0.7mm]
\multicolumn{1}{l|}{$\mathcal{D}_{\textsl{RB}}$} & \small{NA} & $0.54 \pm 0.04$ & \bm{$0.66 \pm 0.03$} \\[0.7mm] \hline
\multicolumn{4}{c}{\textit{Trainable Parameters}}        
\\ \hline
\multicolumn{1}{l|}{-} & $124.3 * 10^6$ & $\approx124.9 * 10^6$ & $\approx5.6 * 10^6$ \\[0.7mm] \hline
\end{tabular}
\caption{This table shows the metrics of our model, averaged over 10 runs, against RoBERTa-based baselines from \citealt{mokhberian-etal-2024-capturing} (cols 1, 2; values as reported). Col. 2 features the configuration $\alpha>0$; see the original paper for details. Along with average values, we report the standard deviation. The highest values are given in bold.}
\label{tab:results}
\vspace{-1em}
\end{table}

\begin{table}[hbt!]
\centering
\small
\begin{tabular}{lccc}
\multicolumn{1}{l|}{\textit{Dataset}}       & \textit{Single-task} & \textit{AE} & \textit{Ours}  \\[0.7mm] \hline
\multicolumn{4}{c}{\textit{Global-level Accuracy}}         \\ \hline
\multicolumn{1}{l|}{$\mathcal{D}_{\textsl{MDA}}$} & $75.65$ & $75.14$ & \bm{$80.11 \pm 0.2$} \\[0.7mm]
\multicolumn{1}{l|}{$\mathcal{D}_{\textsl{HS-Brexit}}$} & $86.77$ & \bm{$87.03$} & $86.49 \pm 0.8$ \\[0.7mm]
\hline
\multicolumn{4}{c}{\textit{Global-level F1}}       
\\ \hline
\multicolumn{1}{l|}{$\mathcal{D}_{\textsl{MDA}}$} & $73.26$ & $73.60$ & \bm{$78.11 \pm 0.2$} \\[0.7mm]
\multicolumn{1}{l|}{$\mathcal{D}_{\textsl{HS-Brexit}}$} & \bm{$64.60$} & $60.36$ & $58.30 \pm 9.8$ \\[0.7mm]
\hline
\multicolumn{4}{c}{\textit{Trainable Parameters}}       
\\ \hline
\multicolumn{1}{l|}{-} & $124.3 * 10^6$ & $\approx125.5 * 10^6$ & $\approx5.6 * 10^6$ \\[0.7mm]
\hline
\end{tabular}
\caption{This table shows the metrics of our model, averaged over 10 runs, against RoBERTa-based baselines from \citealt{deng2023annotate} (cols 1, 2; values as reported; std not reported). Col. 2 features the configuration $E_a$; see the original paper for details. The highest values are given in bold.}
\label{tab:results2}
\vspace{-1em}
\end{table}

In our experiments, the proposed hypernetwork-based architecture demonstrates generally strong performance across all datasets when compared to both AART and AE baselines, while using substantially fewer trainable parameters. Tables \ref{tab:results} and \ref{tab:results2} report the mean and standard deviation over ten runs for each metric suite as defined in Section \ref{sec:evaluation}.
% \footnote{Although we argue that training separate LoRA adapters for each perspective is generally unfeasible, we report a supplemental comparison of our results against these in Section \ref{sec:vslora} in the appendix.}

On the $\mathcal{D}_{\textsl{MDA}}$ dataset, our model achieves an annotator-level F1 of $70.24 \pm 0.9$ and a global-level F1 of $78.11 \pm 0.2$, thus surpassing both the AART baseline (annotator F1 $69.72 \pm 1.1$, global F1 $77.38 \pm 0.4$) and the AE baseline (global accuracy $75.14$, global F1 $73.60$). This result indicates that our architecture is effective at capturing individual annotator tendencies and producing coherent overall predictions in this scenario.

For $\mathcal{D}_{\textsl{RB}}$, our approach again outperforms AART in annotator-level performance ($73.81 \pm 2.0$ vs.\ $71.10 \pm 3.2$) and yields a competitive global-level F1 of $76.17 \pm 1.8$ (AART: $79.96 \pm 1.9$), suggesting that our hypernetwork is capable of modeling both base and rare annotator behaviors even when the dataset exhibits varied disagreement patterns.

Interestingly, on the more challenging $\mathcal{D}_{\textsl{EPIC}}$ dataset, our model gives lower annotator-level F1 ($53.16 \pm 1.6$) and global-level F1 ($65.11 \pm 1.2$) compared to AART’s $59.67 \pm 0.9$ and $66.16 \pm 1.4$, respectively. We believe this shortcoming stems from the higher complexity of data in EPIC,\footnote{For example, as per \citet{casola2024multipico}, gpt-3.5-turbo yields an F1 of $48.1$ on EPIC in zero-shot settings; this differs from its zero-shot performance on other language understanding tasks, such as sentiment analysis or natural language inference (F1s of $91.13$ and $67.87$ respectively, \citet{ye2023comprehensive}).}
which may require more specialized regularization strategies; still, our model attains a higher item-level disagreement correlation ($0.28 \pm 0.03$ vs.\ $0.20 \pm 0.06$), thus showing that these two perspectivist metrics have a potential discrepancy.

On $\mathcal{D}_{\textsl{HS-Brexit}}$, the comparison with AE yields a global accuracy of $86.49$ (AE: $87.03$) and a global F1 of $58.30$ (AE: $60.36$), which is only slightly below the baseline. Given the dataset size, it translates into a difference of \~8 misclassified instances.

Beyond model generalizability, \texttt{hypernetwork+adapters} solution demonstrates remarkably better parameter efficiency, as it requires only $\approx5.6 * 10^6$ trainable parameters compared to $\approx124.9 * 10^6$ for AART and $\approx125.5 * 10^6$ for AE. This advantage is especially important within the strong perspectivist paradigm, where an ideal model should be able to scale up to hundreds of perspectives and, at the same time, take up a reasonable amount of memory and disk space.

\section{Discussion}
\label{sec:discussion}

Two potentially advantageous aspects of the \texttt{hypernetwork+adapters} setting are (1) keeping the base model's weights intact and (2) adapting the model to all targets by training just one component. In what follows, we analyze why these aspects are especially relevant perspectivist learning.

Preserving the base model's original state could be beneficial for two reasons. Concerning inherently multi-purpose models, such as T5 \cite{raffel2020exploring}, it means that the base model's performance on other tasks will not be affected. This effect is especially important when dealing with large language models: they are often used for a wide range of tasks, and updating their weights can lead to catastrophic forgetting \cite{kirkpatrick2017overcoming} or degradation of their performance on previously learned tasks. For instance, if a model is fine-tuned for a particular task like hate speech detection, preserving the base model's original state ensures that its performance on other tasks remains intact.

Moreover, when adapting a model that has already been fine-tuned for some task irrespective of perspectives, our method can preserve this original, 'perspective-neutral' model. In a realistic setting of personalized content moderation, this model can be used as a fallback option. For example, if some user's view is not covered by the perspectivist model, the original, perspective-neutral model can be used to provide a fallback response.

Another benefit of modeling adapters and keeping the main classifier frozen is the mitigation of potential negative outcomes associated with perspectivist learning. In particular, when a RoBERTa model that is fully trainable is finetuned for perspectivist labels (as is the case in \citealt{deng2023annotate} \& \citealt{mokhberian-etal-2024-capturing}), it is taught to associate the same text with varying labels (one to many). This can cause conflicting gradient steps leading the model away from the optimum. Incorporating annotator features through embeddings is intended to resolve this issue by converting the task back to a one-to-one association. However, it is unclear whether these features receive sufficient weight in the classifier to achieve that. In our framework, there is no discrepancy between inputs and outputs, as the only trainable component is the hypernetwork, which learns a one-to-one correspondence between annotators and their respective adapters. This approach does not expose the base model's weights to controversial targets and thus avoids the associated difficulties.

\begin{table}[hbt!]
\centering
\small
\begin{tabular}{lccc}
\multicolumn{1}{l|}{\textit{Dataset}} & \textit{RoBERTa} & \textit{LoRA} & \textit{Ours}  \\[0.7mm] \hline
\multicolumn{4}{c}{\textit{Trainable Parameters}}       
\\ \hline
\multicolumn{1}{l|}{$\mathcal{D}_{\textsl{MDA}}$} & $334*125 * 10^6$ & $334 * 6.6 * 10^5$ & $5.6 * 10^6$ \\[0.7mm]
\multicolumn{1}{l|}{$\mathcal{D}_{\textsl{EPIC}}$} & $74*125 * 10^6$ & $74 * 6.6 * 10^5$ & $5.4 * 10^6$ \\[0.7mm]
\multicolumn{1}{l|}{$\mathcal{D}_{\textsl{RB}}$} & $819*125 * 10^6$ & $819 * 6.6 * 10^5$ & $5.9 * 10^6$ \\[0.7mm]
\multicolumn{1}{l|}{$\mathcal{D}_{\textsl{HS-Brexit}}$} & $6 *125 * 10^6$ & $6 * 6.6 * 10^5$ & $5.3 * 10^6$ \\[0.7mm]
\hline
\end{tabular}
\caption{Overview of trainable parameter counts required for fine-tuning a RoBERTa model to each perspective using all parameters, low-rank adaptation, and our hypernetwork respectively ($r=2,\alpha=32$).}
\label{tab:vsLoRA}
\vspace{-1em}
\end{table}

Using a hypernetwork to learn adapter weights offers an additional advantage. Training a separate LoRA-style adapter for each annotator is possible but involves training a large number of parameters that grows substantially with each added perspective. As shown in Table \ref{tab:vsLoRA}, the costs of adapting RoBERTa-base to every perspective in our datasets are substantial. In extreme cases, such as $\mathcal{D}_{\textsl{RB}}$ with 819 annotators, the required parameters exceed not only the hypernetwork's but also RoBERTa's total parameter count. This results in impractical memory and disk space requirements, rendering the separate adapter approach infeasible. 

In contrast, our method stands out as the more cost-effective solution in all settings, except for $\mathcal{D}_{\textsl{HS-Brexit}}$, which has an exceptionally small number of annotators\footnote{In order to verify that separately-trained LoRA adapters do not significantly outperform our model, we conduct an experiment on $\mathcal{D}_{\textsl{HS-Brexit}}$. We report the outcomes in Table \ref{tab:results3}; they suggest that training separate adapters does not surpass our approach.}. Like other approaches that make use of annotator embeddings, adding a new annotator requires $1 \times hdim$ parameters + rescaling the model to a new embedding. Given the small size of the hypernetwork, all of this sums up to just $2 \times hdim$, allowing for very cheap scaling to additional perspectives. This property makes it especially attractive for perspectivist learning.

\section*{Conclusion}

In this work, we have investigated a \texttt{hypernetwork+adapters} architecture for perspectivist learning on subjective classification tasks. Our experiments show that, while this model does not uniformly exceed the latest perspective‐aware baselines, it achieves superior performance on several perspectivist datasets, most notably $\mathcal{D}_{\textsl{MDA}}$ and $\mathcal{D}_{\textsl{RB}}$. It also obtains higher item‐level disagreement correlations even when mean F1 is lower, as on $\mathcal{D}_{\textsl{EPIC}}$. Notably, our approach requires only about $5.6\times 10^6$ trainable parameters, about 4.5\% with respect to over $124\times 10^6$ in the competing methods; thus, it offers a considerable advantage in parameter efficiency. 

Taken together, these findings suggest that the \texttt{hypernetwork+adapters} design is a promising solution within the strong perspectivist paradigm, even if further work is needed to let it scale equally well to all available tasks and datasets.

\section*{Limitations}
\label{sec:limitations}

One important limitation of the proposed hypernetwork architecture is that it takes more time for inference, as it follows a two-stage procedure where it first separately predicts the LoRA weights of each adapted layer and then applies these during classification. This flow leads to both training and evaluation taking longer than in the case of regular classifier models ($\approx0.25$ iterations/sec. vs. 4 iterations/sec.). We argue, however, that this shortcoming does not outweigh the advantage in parameter efficiency. The decreased memory consumption allows for more parallel training jobs to be scheduled simultaneously, thus compensating for lower throughput.

A further possible limitation is that we only use annotator IDs as annotator features. This strategy does not permit our system to scale to new perspectives if we need to make predictions for an unseen annotator. We see two ways to address this limitation. First, in this paper, we do not inquire into how well hypernetworks work with sociodemographic features. However, they can still be trivially integrated, possibly mitigating this issue. A further way to tackle this problem could consist in finding an annotator in the existing annotator pool that is most similar to the unseen one and assuming their perspective. Similarity of annotators could be approximated from the said sociodemographic features.

Finally, we acknowledge that our evaluation of the proposed architecture is not exhaustive, and we could overlook some shortcomings of our model. However, we find it sufficient to judge how well it compares to the competitor models. Additionally, we hope to scrutinize it more when more perspectivist models and datasets are released.

\section*{Acknowledgments}

We thank the anonymous reviewers and the NLP group at the University of Utrecht for their helpful feedback on this work. Generative tools, such as LLAMA-3 \cite{grattafiori2024llama}, were used to proof-read this text. This study is funded by NWO through an AINed Fellowship Grant NGF.1607.22.002 and supported by project ‘Dealing with Meaning Variation in NLP’.

% initializations of the base model
% However, it appeared to be independent of 
% We hypothesize that this setback may be due to the trainable component of our architecture (i.e., the hypernetwork) having fewer layers compared to deeper networks and thus being more susceptible to bias towards the majority class. A similar tendency was noted by TODO in a controlled experiment that deeper MLP architectures are less likely to be affected by data imbalance.
% On the other hand, irony classification and EPIC data in particular present a difficult task even for full-fledged RoBERTa fine-tuning, as shown by AART. This could result in two different kinds of issues for our architecture: 1) the number of hypernetwork parameters may be insufficient to capture the complexity of the target variable; 2) the predicted adapter parameters are too few to efficiently adapt the model. If this is the case, the problem can be alleviated by trivially growing the number of parameters

\bibliography{paper}

\appendix
\clearpage
\section{Separate LoRA training.}

\begin{table}[hbt!]
\centering
\small
\begin{tabular}{lccc}
\multicolumn{1}{l|}{\textit{Dataset}}       & \textit{Baseline} & \textit{LoRA} & \textit{Ours}  \\[0.7mm] \hline
\multicolumn{4}{c}{\textit{Global-level Accuracy}}         \\ \hline
\multicolumn{1}{l|}{$\mathcal{D}_{\textsl{HS-Brexit}}$} & $86.77$ & $77.50$ & $86.49$ \\
\hline
\multicolumn{4}{c}{\textit{Global-level F1}}       
\\ \hline
\multicolumn{1}{l|}{$\mathcal{D}_{\textsl{HS-Brexit}}$} & $64.60$ & $53.02$ & $58.30$ \\
\hline
\end{tabular}
\caption{This table reports the metrics of our model against the performance of 6 separate LoRA adapters (one per each annotator perspective) on HS-Brexit. The training parameters for the adapters copied those of the main experiment, the exception being an increased learning rate ($5\mathrm{e}{-5}$) and an increased epoch count (10). We report the mean results per 10 runs.}
\label{tab:results3}
\vspace{-1em}
\end{table}

\end{document}